\documentclass[10pt,twocolumn,letterpaper]{article}

\pdfoutput=1

\usepackage{cvpr}
\usepackage{times}
\usepackage{epsfig}
\usepackage{graphicx}
\usepackage{amsmath}
\usepackage{amssymb}
\usepackage{caption}
\usepackage{subcaption}
\usepackage{adjustbox}
\usepackage{multicol,bm,cite}

\newcommand{\R}{{\mathbb{R}}}

\newcommand\mypara[1]{\vspace{0.2cm}\noindent \textbf{#1} \hspace{0.2cm}}

\cvprfinalcopy 

\def\cvprPaperID{1444} 

\title{InverseRenderNet: Learning single image inverse rendering}

\author{Ye Yu and William A. P. Smith\\
Department of Computer Science, University of York, UK\\
{\tt\small \{yy1571,william.smith\}@york.ac.uk}
}

\begin{document}
\twocolumn[{%
\renewcommand\twocolumn[1][]{#1}%
\maketitle
\begin{center}
    \centering
    \begingroup
\setlength{\tabcolsep}{1pt}
\renewcommand{\arraystretch}{0.5}
\begin{tabular}{ccccccc}
\includegraphics[width=2.4cm]{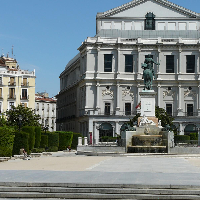}&
\includegraphics[width=2.4cm]{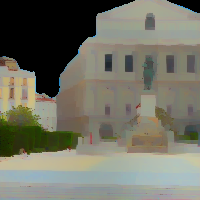}&
\includegraphics[width=2.4cm]{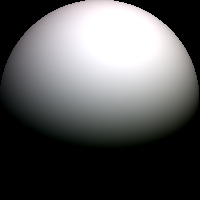}&
\includegraphics[width=2.4cm]{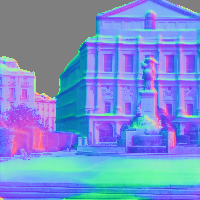}&
\includegraphics[width=2.4cm]{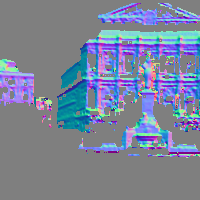}&
\includegraphics[width=2.4cm]{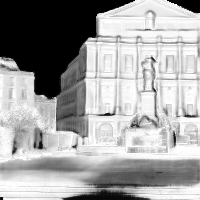}&
\includegraphics[width=2.4cm]{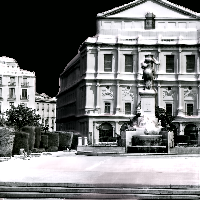}\\
\includegraphics[width=2.4cm]{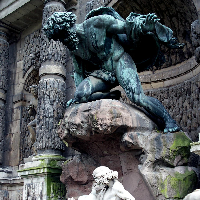}&
\includegraphics[width=2.4cm]{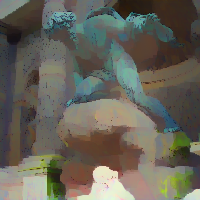}&
\includegraphics[width=2.4cm]{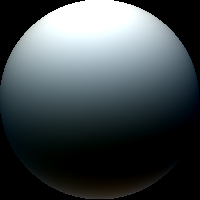}&
\includegraphics[width=2.4cm]{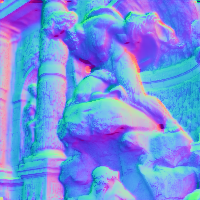}&
\includegraphics[width=2.4cm]{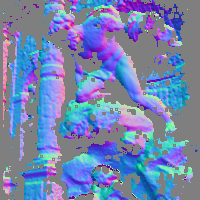}&
\includegraphics[width=2.4cm]{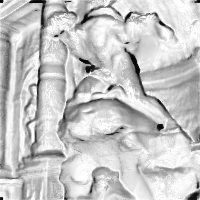}&
\includegraphics[width=2.4cm]{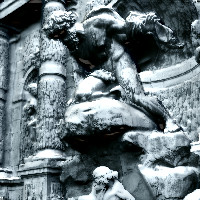}\\
\includegraphics[width=2.4cm]{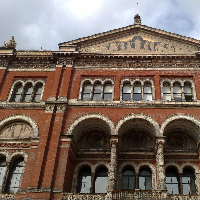}&
\includegraphics[width=2.4cm]{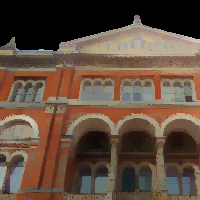}&
\includegraphics[width=2.4cm]{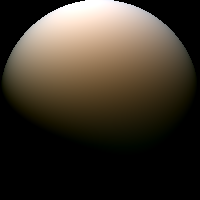}&
\includegraphics[width=2.4cm]{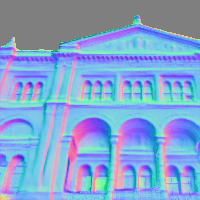}&
\includegraphics[width=2.4cm]{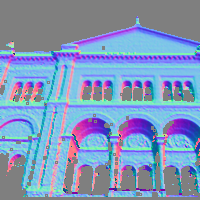}&
\includegraphics[width=2.4cm]{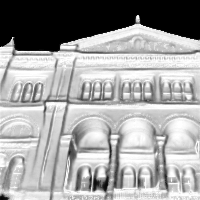}&
\includegraphics[width=2.4cm]{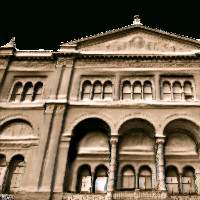}\\
Input & Diffuse albedo & Illumination & NM prediction & NM from MVS & Frontal shading & Shading \\
\end{tabular}
\endgroup
    \vspace{-0.2cm}
    \captionof{figure}{From a single image (col.~1), we estimate albedo and normal maps and illumination (col.~2-4); comparison multiview stereo result from several hundred images (col.~5); re-rendering of our shape with frontal/estimated lighting (col.~6-7).}
    \label{fig:teaser}
\end{center}%
}]



\begin{abstract}
We show how to train a fully convolutional neural network to perform inverse rendering from a single, uncontrolled image. The network takes an RGB image as input, regresses albedo and normal maps from which we compute lighting coefficients. Our network is trained using large uncontrolled image collections without ground truth. By incorporating a differentiable renderer, our network can learn from self-supervision. Since the problem is ill-posed we introduce additional supervision: 1.~We learn a statistical natural illumination prior, 2.~Our key insight is to perform offline multiview stereo (MVS) on images containing rich illumination variation. From the MVS pose and depth maps, we can cross project between overlapping views such that Siamese training can be used to ensure consistent estimation of photometric invariants. MVS depth also provides direct coarse supervision for normal map estimation. We believe this is the first attempt to use MVS supervision for learning inverse rendering.
\end{abstract}

\section{Introduction}\label{sec:intro}

Inverse rendering is the problem of estimating one or more of illumination, reflectance properties and shape from observed appearance (i.e.~one or more images). In this paper, we tackle the most challenging setting of this problem; we seek to estimate all three quantities from only a single, uncontrolled image. Specifically, we estimate a normal map, diffuse albedo map and spherical harmonic lighting coefficients. This subsumes two classical computer vision problems: (uncalibrated) shape-from-shading and intrinsic image decomposition. 

\begin{figure*}[t]
    \centering
    \includegraphics[width=0.8\linewidth,clip=true,trim=18px 278px 57px 321px]{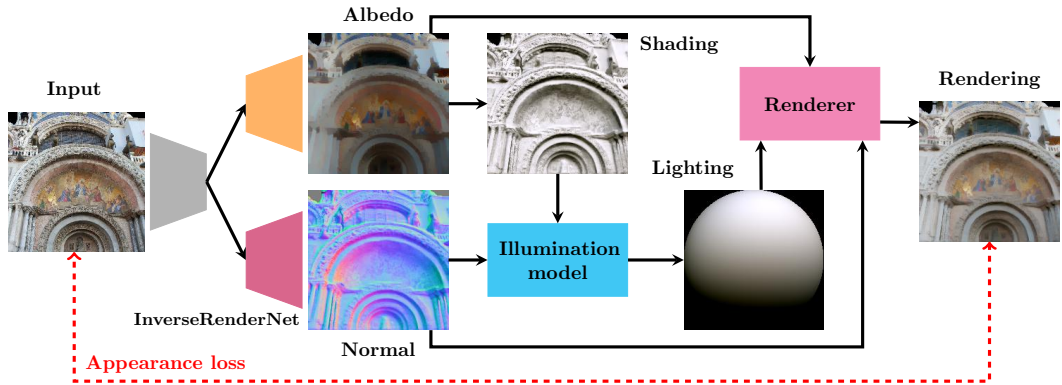}
    \caption{At inference time, our network regresses diffuse albedo and normal maps from a single, uncontrolled image and then computes least squares optimal spherical harmonic lighting coefficients. At training time, we introduce self-supervision via an appearance loss computed using a differentiable renderer and the estimated quantities.}
    \label{fig:inference}
\end{figure*}

Classical approaches \cite{BarronTPAMI2015,Langguth:12} cast these problems in terms of energy minimisation. Here, a data term measures the difference between the input image and the synthesised image that arises from the estimated quantities. We approach the problem as one of image to image translation and solve it using a deep, fully convolutional neural network. However, inverse rendering of uncontrolled, outdoor scenes is itself an unsolved problem and so labels for supervised learning are not available. Instead, we use the data term for self-supervision via a differentiable renderer (see Fig.~\ref{fig:inference}).

Single image inverse rendering is an inherently ambiguous problem. For example, any image can be explained with zero data error by setting the albedo map equal to the image, the normal map to be planar and the illumination arbitrarily such that the shading is unity everywhere. Hence, the data term alone cannot be used to solve this problem. For this reason, classical methods augment the data term with generic \cite{BarronTPAMI2015} or object-class-specific \cite{aldrian2013inverse} priors. Likewise, we also exploit priors during learning (specifically a statistical prior on lighting and a smoothness prior on diffuse albedo). However, our key insight that enables the CNN to learn good performance is to introduce additional supervision provided by an offline multiview reconstruction.

While photometric vision has largely been confined to restrictive lab settings, classical geometric methods are sufficiently robust to provide multiview 3D shape reconstructions from large, unstructured datasets containing very rich illumination variation \cite{Snavely:08, Furukawa:10}. This is made possible by local image descriptors that are largely invariant to illumination. However, these methods recover only geometric information and any recovered texture map has illumination ``baked in'' and so is useless for relighting.
We exploit the robustness of geometric methods to varying illumination to supervise our inverse rendering network. We apply a multiview stereo (MVS) pipeline to large sets of images of the same scene. We select pairs of overlapping images with different illumination, use the estimated relative pose and depth maps to cross project photometric invariants between views and use this for supervision via Siamese training. In other words, geometry provides correspondence that allows us to simulate varying illumination from a fixed viewpoint. Finally, the depth maps from MVS provide coarse normal map estimates that can be used for direct supervision of the normal map estimation.

\subsection{Contribution}

Deep learning has already shown good performance on components of the inverse rendering problem. This includes monocular depth estimation \cite{eigen2014depth}, depth and normal estimation \cite{eigen2015predicting} and intrinsic image decomposition \cite{lettry2018darn}. However, these works use supervised learning. For tasks where ground truth does not exist, such approaches must either train on synthetic data (in which case generalisation to the real world is not guaranteed) or generate pseudo ground truth using an existing method (in which case the network is just learning to replicate the performance of the existing method). Inverse rendering of outdoor, complex scenes is itself an unsolved problem and so reliable ground truth is not available and supervised learning cannot be used. In this context, we make the following contributions. 
To the best of our knowledge, we are the first to exploit MVS supervision for learning inverse rendering. Second, we are the first to tackle the most general version of the problem, considering arbitrary outdoor scenes and learning from real data, as opposed to restricting to a single object class \cite{tewari2017mofa} or using synthetic training data \cite{zheng2018t2net}. Third, we introduce a statistical model of spherical harmonic lighting in natural scenes that we use as a prior. Finally, the resulting network is the first to inverse render all of shape, reflectance and lighting in the wild and we perform the first evaluation in this setting.

\section{Related work}

\paragraph{Classical approaches} 
Classical methods estimate intrinsic properties by fitting photometric or geometric models.
Most methods require multiple images. From multiview images, a structure-from-motion/multiview stereo pipeline enables recovery of dense mesh models \cite{Kazhdan:13, Furukawa:10} though illumination effects are baked into the texture. From images with fixed viewpoint but varying illumination photometric stereo can be applied. Variants consider statistical BRDF models \cite{Alldrin:08}, the use of outdoor time-lapse images \cite{Langguth:12}, spatially-varying BRDFs \cite{Goldman:09}
Attempts to combine geometric and photometric methods are limited. Haber \etal \cite{Haber:09} assume known geometry (which can be provided by MVS) and inverse render reflectance and lighting from community photo collections. Kim \etal \cite{kim2016multi} represents the state-of-the-art and again uses an MVS initialisation for joint optimisation of geometry, illumination and albedo.
Some methods consider a single image setting.
Jeson \etal \cite{jeon2014intrinsic} introduce a local-adaptive reflectance smoothness constraint for intrinsic image decomposition on texture-free input images which are acquired with a texture separation algorithm. Barron \etal \cite{BarronTPAMI2015} present SIRFS, a classical optimisation-based approach that recovers all of shape, illumination and albedo using a sophisticated combination of generic priors.

\mypara{Deep depth prediction}
Direct estimation of shape alone using deep neural networks has attracted a lot of attention. Eigen \etal \cite{eigen2014depth, eigen2015predicting} were the first to apply deep learning in this context. Subsequently, performance gains were obtained using improved architectures \cite{laina2016deeper}, post-processing with classical CRF-based methods \cite{wang2015towards, liu2015deep, xu2017multi} and using ordinal relationships for objects within the scenes \cite{fu2018deep, MegaDepthLi18, chen2016single}. Zheng \etal \cite{zheng2018t2net} use synthetic images for training but improve generalisation using a synthetic-to-real transform GAN. However, all of this work requires supervision by ground truth depth. An alternative branch of methods explore using self-supervision from augmented data. For example, binocular stereo pairs can provide a supervisory signal through consistency of cross projected images \cite{ummenhofer2017demon, kendall2017end, garg2016unsupervised, godard2017unsupervised}. Alternatively, video data can provide a similar source of supervision \cite{zhou2017unsupervised, vijayanarasimhan2017sfm, wang2018learning}. Some of other work built from specific ways were proposed recently. Tulsiani \etal \cite{tulsiani2017multi} use multiview supervision in a ray tracing network. While all these methods take single image input, Ji \etal \cite{ji2017surfacenet} tackle the MVS problem itself using deep learning.  

\mypara{Deep intrinsic image decomposition}
Intrinsic image decomposition is a partial step towards inverse rendering. It decomposes an image into reflectance (albedo) and shading but does not separate shading into shape and illumination. Even so, the lack of ground truth training data makes this a hard problem to solve with deep learning. Recent work either uses synthetic training data and supervised learning \cite{narihira2015direct,han2018learning, lettry2018darn, e.20181172,fan2017revisiting} or self-supervision/unsupervised learning. Very recently, Li \etal \cite{BigTimeLi18} used uncontrolled time-lapse images allowing them to combine an image reconstruction loss with reflectance consistency between frames. This work was further extended using photorealistic, synthetic training data \cite{li2018cgintrinsics}. Ma \etal \cite{ma2018single} also trained on time-lapse sequences and introduced a new gradient constraint which encourage better explanations for sharp changes caused by shading or reflectance. Baslamisli \etal \cite{Baslamisli_2018_CVPR} applied a similar gradient constraint while they used supervised training. Shelhamer \etal \cite{shelhamer2015scene} propose a hybrid approach where a CNN estimates a depth map which is used to constrain a classical optimisation-based intrinsic image estimation.

\mypara{Deep inverse rendering}
To date, this topic has not received much attention. One line of work simplifies the problem by restricting to a single object class, e.g.~faces \cite{tewari2017mofa}, meaning that a statistical face model can constrain the geometry and reflectance estimates. This enables entirely self-supervised training. Shu \etal \cite{shu2017neural} extend this idea with an adversarial loss. Sengupta \etal \cite{sengupta2017sfsnet} on the other hand, initialise with supervised training on synthetic data, and fine-tuned their network in an unsupervised fashion on real images. Another line of work restricts geometry to almost planar objects and lighting to a flash in the viewing direction  \cite{aittala2016reflectance,li2017modeling} under which assumptions they can obtain impressive results.
More general settings have been considered by Kulkarni \etal \cite{kulkarni2015deep} who show how to learn latent variables that correspond to extrinsic parameters allowing image manipulation. Janner \etal \cite{janner2017self} is the only prior work we are aware of that tackles the full inverse rendering problem. Like us, they use self-supervision but include a trainable shading model. However, the shader requires supervised training on synthetic data, limiting the ability of the network to generalise to real world scenes.

\section{Preliminaries}


We assume that a perspective camera observes a scene, such that the projection from 3D world coordinates, $(u,v,w)$, to 2D image coordinates, $(x,y)$, is given by:
\small
\begin{equation}
    \lambda \begin{bmatrix}x\\y\\1\end{bmatrix} = 
    \mathbf{P}
    \begin{bmatrix}u\\v\\w\\1\end{bmatrix},\quad \mathbf{P} = \mathbf{K}
    \begin{bmatrix}\mathbf{R} & \mathbf{t}\end{bmatrix},\quad \mathbf{K} = \begin{bmatrix}f & 0 & c_x \\ 0 & f & c_y \\ 0 & 0 & 1\end{bmatrix},
\end{equation}
\normalsize
where $\lambda$ is an arbitrary scale factor, $\mathbf{R}\in SO(3)$ a rotation matrix, $\mathbf{t}\in\R^3$ a translation vector, $f$ the focal length and $(c_x,c_y)$ the principal point. 

The inverse rendered shape estimate could be represented in a number of ways. For example, many previous methods estimate a viewer-centred depth map. However, local reflectance, and hence appearance, is determined by surface orientation, i.e.~the local surface normal direction. So, to render a depth map for self-supervision, we would need to compute the surface normal. From a perspective depth map $w(x,y)$, the surface normal direction is:
\begin{equation}
    \bar{\mathbf{n}} = \begin{bmatrix}-fw_x(x,y)\\-fw_y(x,y)\\(x-c_x)w_x(x,y) + (y-c_y)w_y(x,y) + w(x,y)\end{bmatrix}\label{eqn:perspnorm}
\end{equation}
from which the unit length normal is given by:
$    \mathbf{n}=\bar{\mathbf{n}}/\|\bar{\mathbf{n}}\|
$.
The derivatives of the depth map in the image plane, $w_x(x,y)$ and $w_y(x,y)$, can be approximated by finite differences. However, \eqref{eqn:perspnorm} requires knowledge of the intrinsic camera parameters. This would severely restrict the applicability of our method. For this reason, we choose to estimate a surface normal map directly.

Although the surface normal can be represented by a 3D vector, since $\|\mathbf{n}\|=1$ it has only two degrees of freedom. So, our network estimates the two elements of the surface gradient at each pixel, $w_u(x,y)$ and $w_u(x,y)$, and the transformation to a 3D surface normal vector is computed by a fixed layer that calculates: $\bar{\mathbf{n}} = [-w_u(x,y),-w_u(x,y),1]^T$. Note that we estimate the normal map in a viewer-centred coordinate system.

We assume that appearance can be approximated by a local reflectance model under environment illumination. Specifically we use a Lambertian diffuse model with order 2 spherical harmonic lighting. This means that RGB intensity can be computed as
\begin{equation}
    \mathbf{i}_{\textrm{lin}}(\mathbf{n},\bm{\alpha},\mathbf{L}) = \textrm{diag}(\bm{\alpha})\mathbf{L}\mathbf{b}(\mathbf{n}),\label{eqn:imageform}
\end{equation}
where $\mathbf{L}\in\R^{3\times 9}$ contains the spherical harmonic colour illumination coefficients, $\bm{\alpha}=[\alpha_r,\alpha_g,\alpha_b]^T$ is the colour diffuse albedo and the order 2 basis is given by:
{\small \begin{equation}
    \mathbf{b}(\mathbf{n})=[1,n_x,n_y,n_z,3n_z^2-1,n_xn_y,n_xn_z,n_yn_z,n_x^2-n_y^2]^T.\label{eqn:basis}
\end{equation}}
Our appearance model means that we neglect high frequency illumination effects, cast shadows and interreflections. However, we found that in practice this model works well for typical outdoor scenes. Finally, cameras apply a nonlinear gamma transformation. We simulate this to produce our final predicted intensities:
$
    \mathbf{i}_{\textrm{pred}}=\mathbf{i}_{\textrm{lin}}^{1/\gamma}
$,
where we assume a fixed $\gamma=2.2$.

\section{Architecture}

Our inverse rendering network (see Fig.~\ref{fig:inference}) is an image-to-image network that regresses albedo and normal maps from a single image and uses these to estimate lighting. We describe these inference components in more detail here.

\subsection{Trainable encoder-decoder}

We implement a deep fully-convolutional neural network with skip connections like the hourglass architecture \cite{newell2016stacked}. We use a single encoder and separate deconvolution decoders for albedo and normal prediction. Albedo maps have 3 channel RGB output, normal maps have two channels for the surface gradient which is converted to a normal map as described above. Both convolutional subnet and deconvolutional subnet contain 15 layers and the activation functions are ReLUs. Adam Optimiser is used in training.

\subsection{Implicit lighting prediction}

In order to estimate illumination parameters, one option would be to use a fully connected branch from the output of our decoder and train our network to predict it directly. However, fully connected layers require very large numbers of parameters and, in fact, lighting can be inferred from the input image and estimated albedo and normal maps, making its explicit prediction redundant. An additional advantage is that the architecture remains fully convolutional and so can process images of any size at inference time.

Consider an input image comprising $K$ pixels. We invert the nonlinear gamma and stack the linearised RGB values to form the matrix $\mathbf{I}\in\R^{3\times K}$. We similarly stack the estimated albedo map to form $\mathbf{A}\in\R^{3\times K}$, the estimated surface normals to form $\mathbf{N}\in\R^{3\times K}$ and define $\mathbf{B}(\mathbf{N})\in\R^{9\times K}$ by applying \eqref{eqn:basis} to each normal vector. We can now rewrite \eqref{eqn:imageform} for the whole image as:
\begin{equation}
    \mathbf{I} = \mathbf{A} \odot \mathbf{L}\mathbf{B}(\mathbf{N}),\label{eqn:illummodim}
\end{equation}
where $\odot$ is the Hadamard (element-wise) product. We can now solve for the spherical harmonic illumination coefficients in a least squares sense, using the whole image. This can be done using any method, so long as the computation is differentiable such that losses dependent on the estimated illumination can have their gradients backpropagated into the inverse rendering network. For example, the solution using the pseudoinverse is given by:
$    \mathbf{L} = (\mathbf{I}\oslash\mathbf{A})\mathbf{B}(\mathbf{N})^+
$,
where $\oslash$ denotes element-wise division and $\mathbf{B}(\mathbf{N})^+$ is the pseudoinverse of $\mathbf{B}(\mathbf{N})$. Fig.~\ref{fig:inference} shows the inferred shading, $\mathbf{I}\oslash\mathbf{A}$, and a visualisation of the estimated lighting.

\section{Supervision}

As shown in Fig.~\ref{fig:inference}, we use a data term (the error between predicted and observed appearance) for self-supervision. However, inverse rendering using only a data term is ill-posed (an infinite set of solutions can yield zero data error) and so we use additional sources of supervision, all of which are essential for good performance. We describe all sources of supervision in this section.

\subsection{Self-supervision via differentiable rendering}

Given estimated normal and albedo maps and spherical harmonic illumination coefficients, we compute a predicted image using \eqref{eqn:imageform}. This local illumination model is straightforward to differentiate. Self-supervision is provided by the error between the predicted, $\mathbf{i}_{\textrm{pred}}$, and observed, $\mathbf{i}_{\textrm{obs}}$, intensities. We compute this error in LAB space as this provides perceptually more convincing results:
\begin{equation}
    \ell_{\textrm{appearance}} = \| \textrm{LAB}(\mathbf{i}_{\textrm{pred}})-\textrm{LAB}(\mathbf{i}_{\textrm{obs}}) \|,\label{eqn:apploss}
\end{equation}
where LAB performs the colour space transformation.

\begin{figure}
    \centering
        \begin{minipage}[c]{.33\linewidth}
            \begin{subfigure}[b]{\linewidth}
                \centering
                \includegraphics[width=0.9\linewidth]{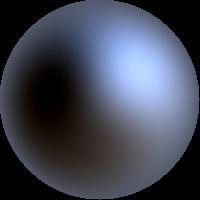}
                \caption{mean\,+\,1st}
            \end{subfigure}\vspace{10pt}
            
            \begin{subfigure}[b]{\linewidth}
                \centering
                \includegraphics[width=0.9\linewidth]{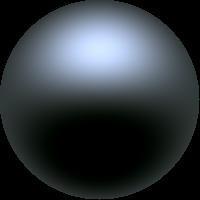}
                \caption{mean\,+\,2nd}
            \end{subfigure}
        \end{minipage}
        \begin{minipage}[c]{.33\linewidth}
            \begin{subfigure}[b]{\linewidth}
                \centering
                \includegraphics[width=0.9\linewidth]{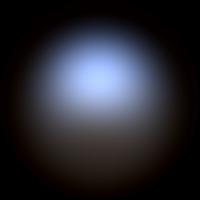}
                \caption{mean\,-\,3rd}
            \end{subfigure}\vspace{-3pt}
            
            \begin{subfigure}[b]{\linewidth}
                \centering
                \includegraphics[width=0.9\linewidth]{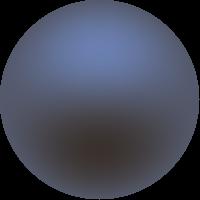}
                \caption{mean}
            \end{subfigure}\vspace{-3pt}
            
            \begin{subfigure}[b]{\linewidth}
                \centering
                \includegraphics[width=0.9\linewidth]{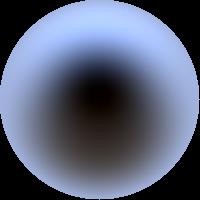}
                \caption{mean\,+\,3rd}
            \end{subfigure}
        \end{minipage}
        \begin{minipage}[c]{.33\linewidth}
            \begin{subfigure}[b]{\linewidth}
                \centering
                \includegraphics[width=0.9\linewidth]{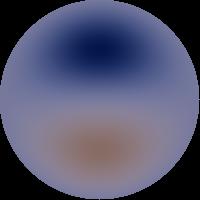}
                \caption{mean\,-\,2nd}
            \end{subfigure}\vspace{10pt}
            
            \begin{subfigure}[b]{\linewidth}
                \centering
                \includegraphics[width=0.9\linewidth]{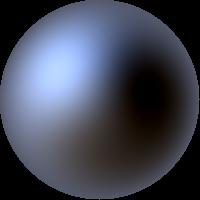}
                \caption{mean\,-\,1st}
            \end{subfigure}
        \end{minipage} 
        \caption{Statistical illumination model. The central image shows the mean illumination. The two diagonals and the vertical show the first 3 principal components.} 
 \label{fig:illu_model}     
 \end{figure}
\subsection{Natural illumination model and prior}

The spherical harmonic lighting model in \eqref{eqn:imageform} enables efficient representation of complex lighting. However, even within this low dimensional space, not all possible illumination environments are natural. The space of natural illumination possesses statistical regularities \cite{dror2001statistics}. We can use this knowledge to constrain the space of possible illumination and enforce a prior on the illumination parameters. To do this, we build a statistical illumination model (see Fig.~\ref{fig:illu_model}) using a dataset of 79 HDR spherical panoramic images taken outdoors. For each environment, we compute the spherical harmonic coefficients, $\mathbf{L}_i\in\R^{3\times 9}$. Since the overall intensity scale is arbitrary, we also normalise each lighting matrix to unit norm, $\|\mathbf{L}_i\|_{\textrm{Fro}}=1$, to avoid ambiguity with the albedo scale. Our illumination model in \eqref{eqn:illummodim} uses surface normals in a viewer-centred coordinate system. So, the dataset must be augmented to account for possible rotations of the environment relative to the viewer. Since the rotation around the vertical ($v$) axis is arbitrary, we rotate the lighting coefficients by angles between $0$ and $2\pi$ in increments of $\pi/18$. In addition, to account for camera pitch or roll, we additionally augment with rotations about the $u$ and $w$ axes in the range $(-\pi/6,\, \pi/6)$. This gives us a dataset of 139,356 environments.
We then build a statistical model, such that any illumination can be approximated as:
\begin{equation}\label{equ:L}
    \textrm{vec}(\mathbf{L}) = \mathbf{P}\textrm{diag}(\sigma_1,\dots,\sigma_D){\bm \alpha} + \textrm{vec}(\mathbf{\bar{L}}).
\end{equation}
where $\mathbf{P}\in\R^{27\times D}$ contains the principal components, $\sigma^2_1,\dots,\sigma^2_D$ are the corresponding eigenvalues, $\mathbf{\bar{L}}\in\R^{3\times 9}$ is the mean lighting coefficients and ${\bm \alpha}\in\R^D$ is the parametric representation of $\mathbf{L}$. We use $D=18$ dimensions. Under the assumption that the original data is Gaussian distributed then the parameters are normally distributed: ${\bm \alpha}\sim\cal{N}(\mathbf{0},\mathbf{I})$. When we compute lighting, we do so within the subspace of the statistical model. In addition, we introduce a prior loss on the estimated lighting vector: $\ell_{\textrm{lighting}} = \| {\bm \alpha} \|^2$.

\begin{figure*}[t]
    \centering
    \includegraphics[width=0.9\linewidth,clip=true,trim=39px 203px 14px 254px]{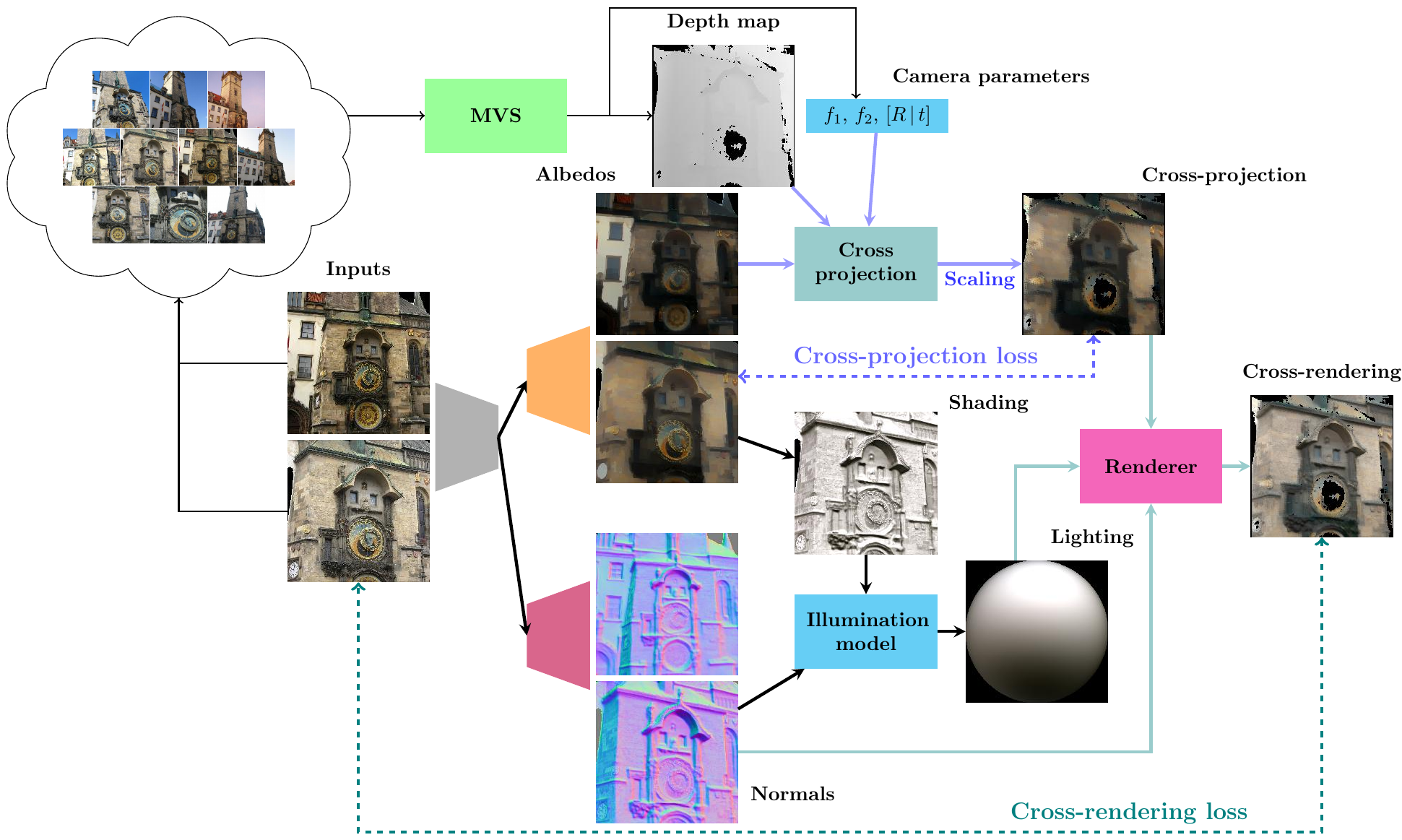}
    \caption{Siamese MVS supervision: albedo cross-projection consistency loss and cross-rendering loss.}
    \label{fig:supervision}
\end{figure*}

\subsection{Multiview stereo supervision}

A pipeline comprising structure-from-motion followed by multiview stereo (which we refer to simply as MVS) enables both camera poses and dense 3D scene models to be estimated from large, uncontrolled image sets. Of particular importance to us, these pipelines are relatively insensitive to illumination variation between images in the dataset since they rely on matching local image features that are themselves illumination insensitive. 
We emphasise that MVS is run offline prior to training and that at inference time our network uses only single images of novel scenes. We use the MVS output for three sources of supervision.

\mypara{Cross-projection}
We use the MVS poses and depth maps to establish correspondence between views, allowing us to cross-project quantities between overlapping images. Given an estimated depth map, $w(x,y)$, in view $i$ and camera matrices for views $i$ and $j$, a pixel $(x,y)$ can be cross-projected to location $(x^{\prime},y^{\prime})$ in view $j$ via:
\begin{equation}
\lambda\begin{bmatrix}
x^{\prime}\\
y^{\prime}\\
1
\end{bmatrix}=
    \mathbf{P}_j\begin{bmatrix}\mathbf{R}_i^T&-\mathbf{R}_i^T\mathbf{t}_i\\ \mathbf{0} & 1\end{bmatrix}
    \begin{bmatrix}
    w(x,y)\mathbf{K}_i^-1\begin{bmatrix}x\\y\\1\end{bmatrix}\\
    1\end{bmatrix}
\end{equation}
In practice, we perform the cross-projection in the reverse direction, computing non-integer pixel locations in the source view for each pixel in the target view. We can then use bilinear interpolation of the source image to compute quantities for each pixel in the target image. Since the MVS depth maps contain holes, any pixels that cross project to a missing pixel are not assigned a value. Similarly, any target pixels that project outside the image bounds of the source are not assigned a value.

\mypara{Direct normal map supervision}
The per-view depth maps provided by MVS can be used to estimate normal maps, albeit that they are typically coarse and incomplete (see Fig.~\ref{fig:teaser}, column 5). We compute guide normal maps from the depth maps and intrinsic camera parameters estimated by MVS using \eqref{eqn:perspnorm}. The guide normal maps are used for direct supervision by computing a loss that measures the angular difference between the guide, $\mathbf{n}_{\textrm{guide}}$, and estimated, $\mathbf{n}_{\textrm{est}}$, surface normals: $\ell_{\textrm{NM}} = \arccos(\mathbf{n}_{\textrm{guide}}\cdot \mathbf{n}_{\textrm{est}})$.

\mypara{Albedo consistency loss}
Diffuse albedo is an intrinsic quantity. Hence, we expect that albedo estimates of the same scene point from two overlapping images should be the same, even if the illumination varies between views. Hence, we automatically select pairs of images that overlap (defined as having similar camera locations and similar centres of mass of their backprojected depth maps). We discard pairs that do not contain illumination variation (where cross-projected appearance is too similar). Then, we train our network in a Siamese fashion on these pairs and use the cross projection described above to compute an albedo consistency loss:
$\ell_{\textrm{albedo}} = \left\| \textrm{LAB}(\mathbf{A}_i) - s\textrm{LAB}(\mathbf{A}_j) \right\|^2_{\textrm{fro}}$,
where $A_i,A_j\in\R^{3\times K}$ are the estimated albedo maps in the $i$th and $j$th images respectively, where $\mathbf{A}_j$ has been cross projected to view $i$, for the $K$ pixels in which image $i$ has a defined depth value. The scalar $s$ is the value that minimises the loss and accounts for the fact that there is an overall scale ambiguity between images. Again, we compute albedo consistency loss in LAB space. The albedo consistency loss is visualised by the blue arrows in Fig.~\ref{fig:supervision}.

\begin{figure*}[!t]
\footnotesize
\centering
\begingroup
\setlength{\tabcolsep}{1pt}
\renewcommand{\arraystretch}{0.5}
\begin{tabular}{ccccccc}
\includegraphics[width=2.4cm]{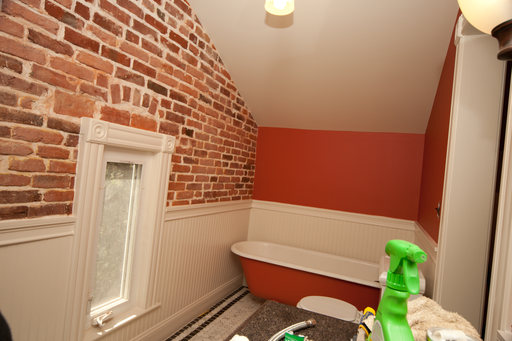}& 
\includegraphics[width=2.4cm]{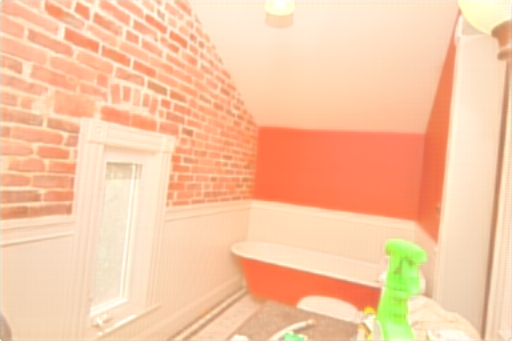}&
\includegraphics[width=2.4cm]{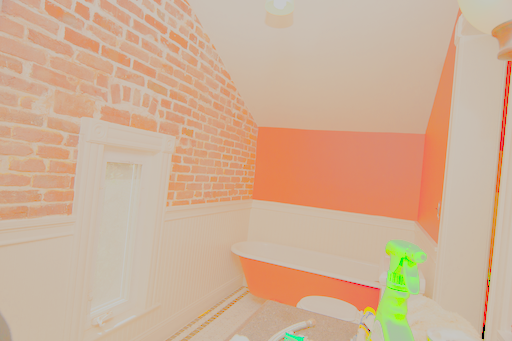}&
\includegraphics[width=2.4cm]{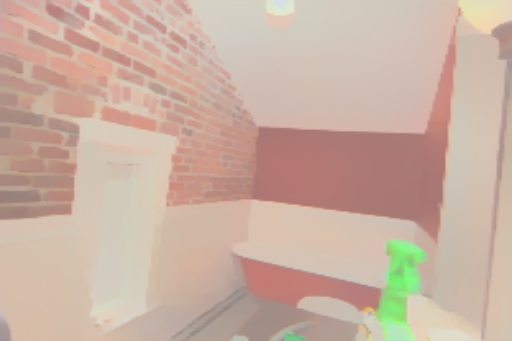}&
\includegraphics[width=2.4cm]{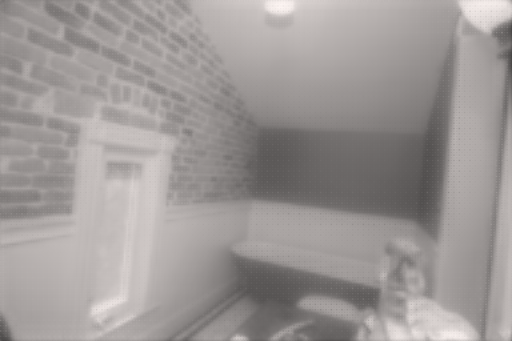}&
\includegraphics[width=2.4cm]{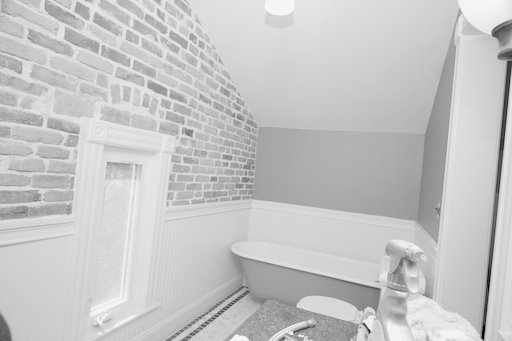}& 
\includegraphics[width=2.4cm]{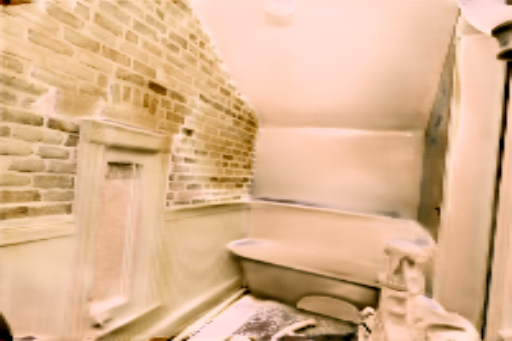}\\
\includegraphics[width=2.4cm]{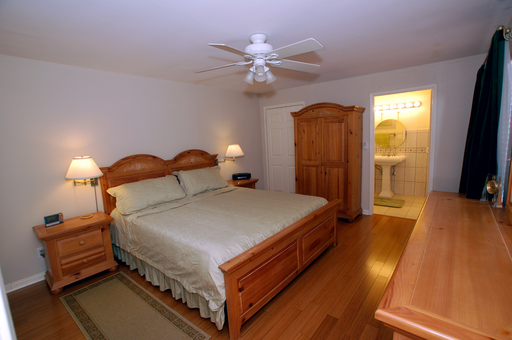}& 
\includegraphics[width=2.4cm]{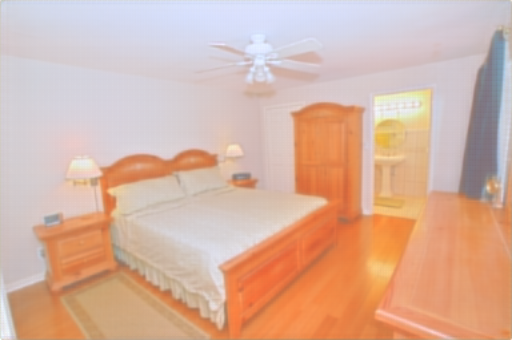}&
\includegraphics[width=2.4cm]{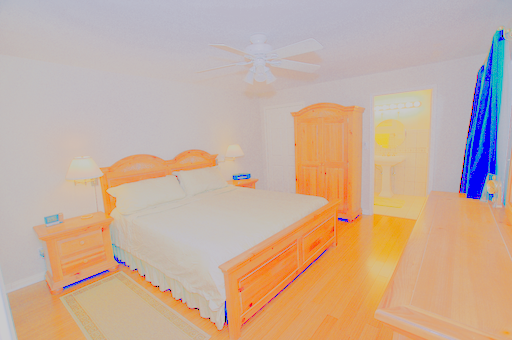}&
\includegraphics[width=2.4cm]{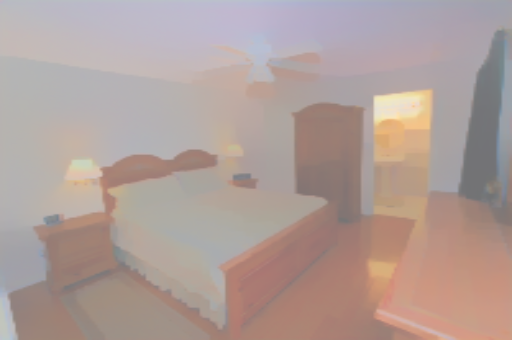}&
\includegraphics[width=2.4cm]{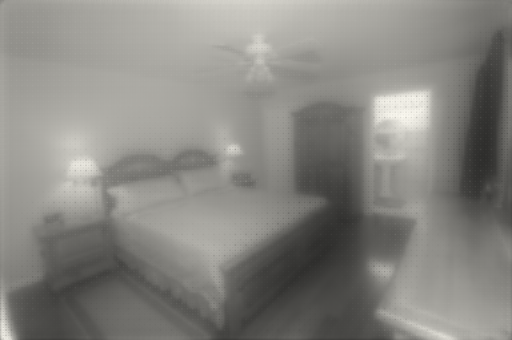}&
\includegraphics[width=2.4cm]{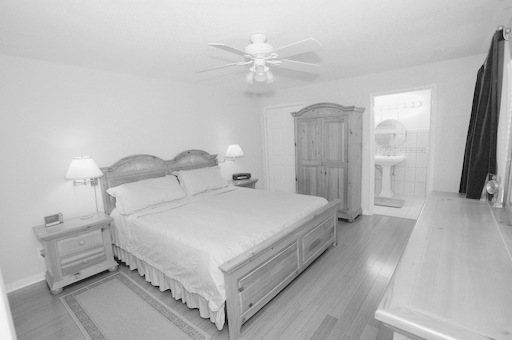}& 
\includegraphics[width=2.4cm]{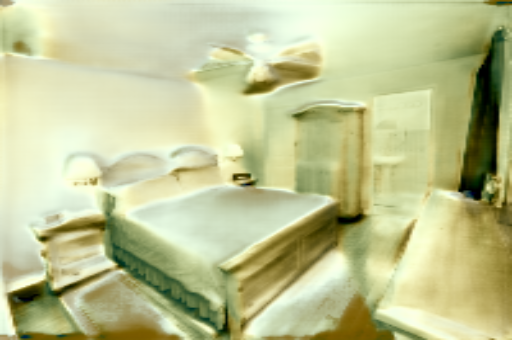} \\
\includegraphics[width=2.4cm]{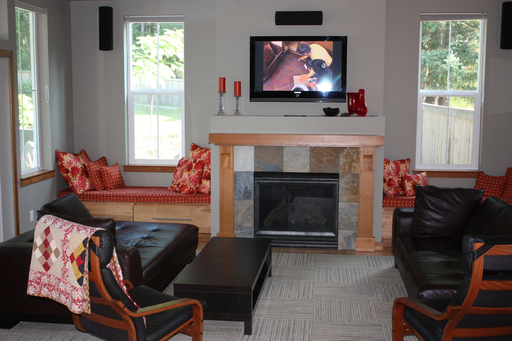}& 
\includegraphics[width=2.4cm]{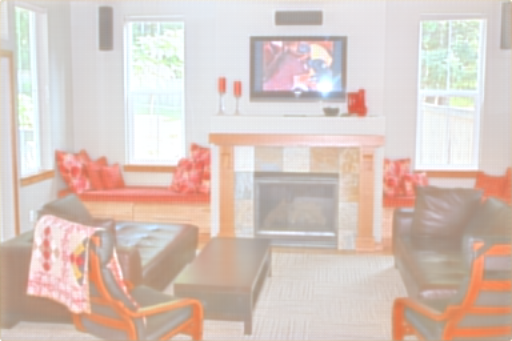}&
\includegraphics[width=2.4cm]{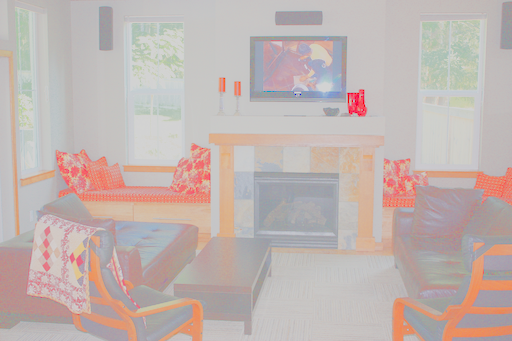}&
\includegraphics[width=2.4cm]{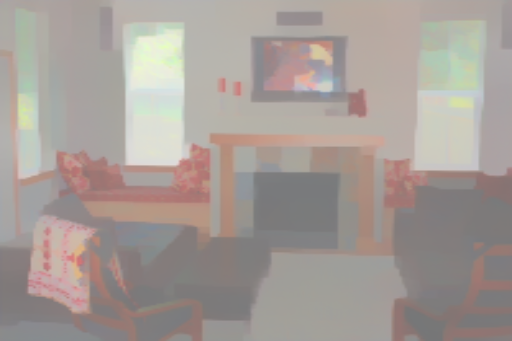}&
\includegraphics[width=2.4cm]{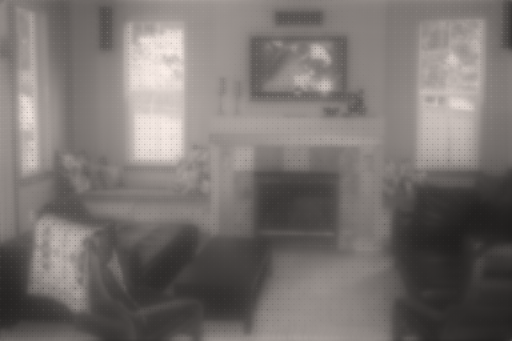}&
\includegraphics[width=2.4cm]{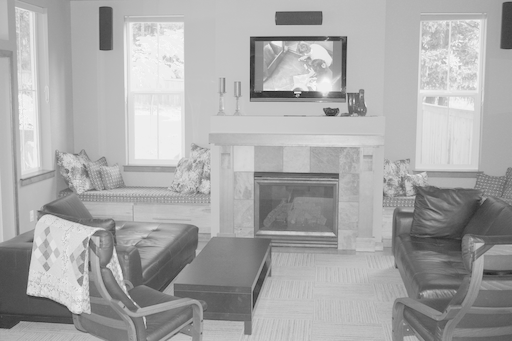}& 
\includegraphics[width=2.4cm]{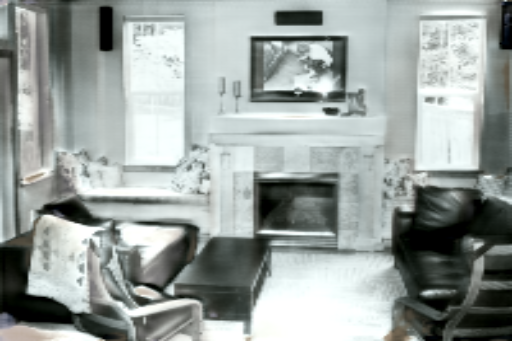} \\
Images & Li \cite{BigTimeLi18} (R) & Nestmeyer \cite{nestmeyer2017reflectance} (R) & Ours (R) & Li \cite{BigTimeLi18} (S) & Nestmeyer \cite{nestmeyer2017reflectance} (S) & Ours (S)\\
\end{tabular}
\endgroup
    \vspace{-0.2cm}
    \caption{Qualitative results for IIW. Second column to forth column are reflectance predictions from \cite{BigTimeLi18}, \cite{nestmeyer2017reflectance} and ours. The last three columns are corresponding shading predictions.}
     \label{fig:iiw_figure}
\end{figure*}%

\mypara{Cross-rendering loss} For improved stability, we also use a mixed cross-projection/appearance loss, $\ell_{\textrm{cross-rend}}$. We use the cross-projected albedo above in conjunction with the estimated normals and albedo to render a new image and measure the appearance error in the same way as \eqref{eqn:apploss}. This loss is visualised by the green arrows in Fig.~\ref{fig:supervision}.

\subsection{Albedo priors}
Finally, we also employ two additional prior losses on the albedo. This helps resolve ambiguities between shading and albedo.
First, we introduce an albedo smoothness prior, $\ell_{\textrm{albedo-smooth}}$. Rather than uniformly applying smoothness penalty, we apply a pixel-wise varying weighted penalty according to chromaticities of the input image. So the stronger smoothness penalties are only enforced on neighbouring pixels with closer chromaticities. The loss itself is the L1 distance between adjacent pixels.

Second, during the self-supervised phase of training, we also introduce a pseudo supervision loss to prevent convergence to trivial solutions. After the pretraining process (see Section 6), our model learns plausible albedo predictions using MVS normals. To prevent subsequent training diverging too far from this, we encourage albedo predictions to remain close to the pretrained albedo predictions. 

\section{Training}

We train our network to minimise:
$\ell = w_1\ell_{\textrm{appearance}} + w_2\ell_{\textrm{NM}} + w_3\ell_{\textrm{albedo}} + w_4\ell_{\textrm{cross-rend}} + w_5\ell_{\textrm{albedo-smooth}} + w_6\ell_{\textrm{albedo-pseudoSup}}$.

\mypara{Datasets} We train using the MegaDepth \cite{MegaDepthLi18} dataset. This contains dense depth maps and camera calibration parameters estimated from crawled Flickr images. The pre-processed images have arbitrary shapes and orientations. For ease of training, we crop square images and resize to a fixed size. We choose our crops to maximise the number of pixels with defined depth values. Where possible, we crop multiple images from each image, achieving augmentation as well as standardisation. We create mini-batches with overlap between all pairs of images in the mini-batch and sufficient illumination variation (correlation coefficient of intensity histograms significantly different from 1). Finally, before inputting an image to our network, we detect and mask the sky region using PSPNet \cite{zhao2017pyramid}. This is because the albedo map and normal map in sky area are meaingless and it severely influences illumination estimation.

\mypara{Training strategy} We found that for convergence to a good solution it is important to include a pre-training phase. During this phase, the surface normals used for illumination estimation and for the appearance-based losses are the MVS normal maps. This means that the surface normal prediction decoder is only learning from the direct supervision loss, i.e. it is learning to replicate the MVS normals. After this initial phase, we switch to full self-supervision where the predicted appearance is computed entirely from estimated quantities. Note that this pre-taining step is not using pseudo albedo supervisions.

\begin{figure*}[!t]
\setlength{\tabcolsep}{1pt}
\renewcommand{\arraystretch}{1.2}
\centering
{
\begin{tabular}{cc}
\includegraphics[width=.5\textwidth,clip=true,trim=75px 60px 62px 15px]{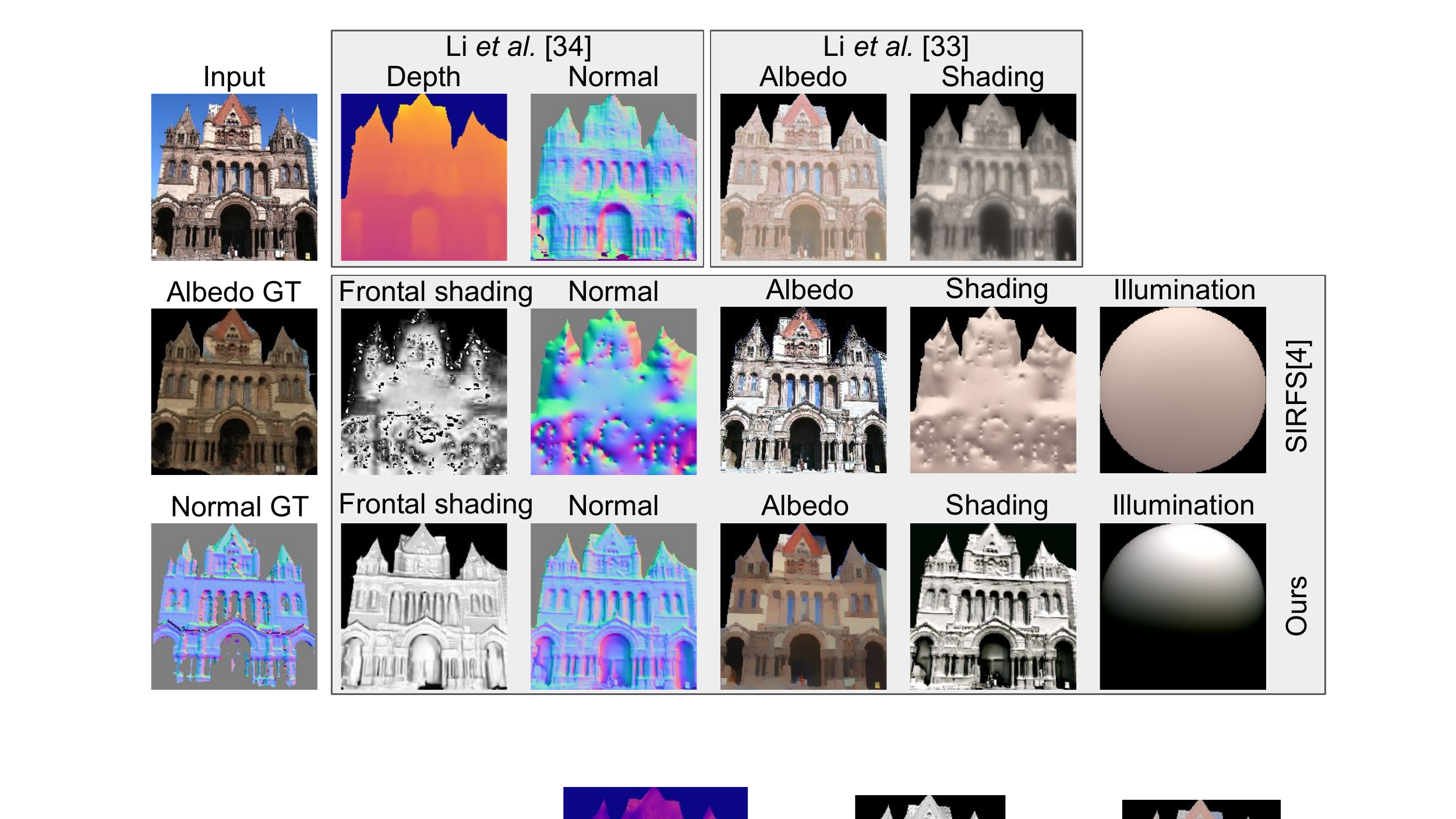} & \includegraphics[width=.5\textwidth,clip=true,trim=75px 60px 62px 15px]{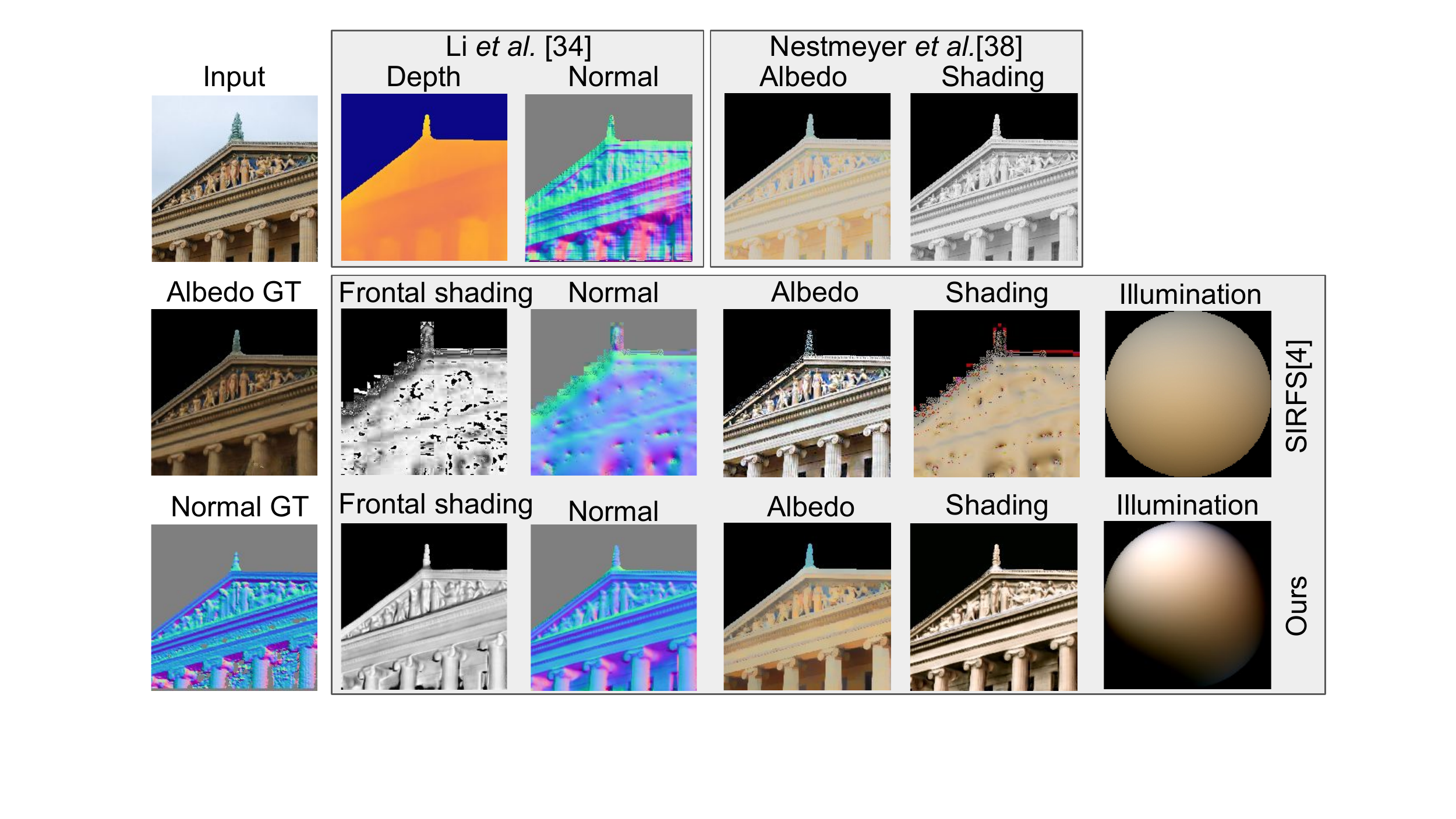} \\
\end{tabular}

    \vspace{-0.2cm}
    \caption{Inverse Rendering Results.}
}
\end{figure*}\label{fig:results}

\section{Evaluation}

There are no existing benchmarks for inverse rendering in the wild. So, we evaluate our method on an intrinsic image benchmark and devise our own benchmark for inverse rendering. Finally, we show a relighting application.

\begin{table}[!t]
\setlength{\tabcolsep}{7pt}
\renewcommand{\arraystretch}{1.1}
\centering
{
    \begin{tabular}{c|c|c}
    \hline
    Methods & Training data &WHDR \\ \cline{1-3}
    \hline
    \hline

    Nestmeyer \cite{nestmeyer2017reflectance} (CNN) & IIW & 19.5 \\
    Zhou \etal \cite{zhou2015learning}  & IIW & 19.9 \\ 
    Fan \etal \cite{fan2017revisiting}  & IIW & 14.5 \\
    DI \cite{narihira2015direct}  & Sintel+MIT & 37.3 \\
    Shi \etal \cite{shi2017learning}  & ShapeNet & 59.4 \\ 
    Li \etal \cite{BigTimeLi18} & BigTime & 20.3 \\ \cline{1-3}
    Ours  & MegaDepth & 21.4 \\ \cline{1-3}
    
    \hline
    \end{tabular}
}
\caption{Evaluation results on IIW benchmark using WHDR percentage (lower is better). The second column shows which dataset on which the networks were trained.}
\label{tab:IIW_table}
\end{table}

\mypara{Evaluation on IIW}
The standard benchmark for intrinsic image decomposition is Intrinsic Images in the Wild \cite{bell14intrinsic} (IIW) which is almost exclusively indoor scenes. Since our training regime requires large multiview image datasets, we are restricted to using scene-tagged images crawled from the web, which are usually outdoors. In addition, our illumination model is learnt on outdoor, natural environments. For these reasons, we cannot perform training or fine-tuning on indoor benchmarks. Moreover, our network is not trained specifically for the task of intrinsic image estimation and our shading predictions are limited by the fact that we use an explicit local illumination model (so cannot predict cast shadows). Nevertheless, we test our network on this benchmark directly without fine-tuning.  We follow the suggestion in \cite{nestmeyer2017reflectance} and rescale albedo predictions to the range $(0.5,1)$ before evaluation. Quantitative results are shown in Tab.~\ref{tab:IIW_table} and some qualitative visual comparison in Fig.~\ref{fig:iiw_figure}. 
Despite the limitations described above, we achieve the second best performance of the methods not trained on the IIW data.

\mypara{Evaluation on MegaDepth}
We evaluate inverse rendering using unobserved scenes from the MegaDepth dataset \cite{MegaDepthLi18}. We evaluate normal estimation performance directly using the MVS geometry. We evaluate albedo estimation using a state-of-the-art multiview inverse rendering algorithm \cite{kim2016multi}. Given the output from their pipeline, we perform rasterisation to generate albedo ground truth for every input image. Note that both sources of ``ground truth'' here are themselves only estimations, e.g.~the albedo ground truth contains ambient occlusion baked in. The colour balance of the estimated albedo is arbitrary, so we compute per-channel optimal scalings prior to computing errors. We use three metrics - MSE, LMSE and DSSIM, which are commonly used for evaluating albedo predictions. To evaluate normal predictions, we use angular errors. The correctness of illumination predictions could be inferred by the other two, so we do not perform explicit evaluations on it. The quantitative evaluations are shown in Tab.~\ref{tab:am&nm_table}. For depth prediction methods, we first compute the optimal scaling onto the ground truth geometry, then differentiate to compute surface normals. These methods can only be evaluated on normal prediction. Intrinsic image methods can only be evaluated on albedo prediction. We can see that our network performs best in normal prediction and also the best in MSE and DSSIM. Qualitative example results can be seen in Fig.~\ref{fig:results}. 

\begin{table}[!t]
\centering
\resizebox{\columnwidth}{!}{%
\setlength{\tabcolsep}{2pt}
\renewcommand{\arraystretch}{1}
    \begin{tabular}{c|ccc|cc}
    \hline
     & \multicolumn{3}{c|}{Reflectances} & \multicolumn{2}{c} {Normals} \\ 
    Methods  & MSE & LMSE & DSSIM & Mean & Median \\
    \hline
    \hline
    Li \etal \cite{MegaDepthLi18}  & - & - & - & 50.6 & 50.4 \\
    Godard \etal \cite{godard2017unsupervised}  & - & - & - & 79.2 & 79.6 \\
    Nestmeyer \etal \cite{nestmeyer2017reflectance}  & 0.0204 & 0.0735 & 0.241 & - & - \\
    Li \etal \cite{BigTimeLi18}  & 0.0171 & 0.0637 & 0.208 & - & - \\
    SIRFS \cite{BarronTPAMI2015} & 0.0383 & 0.222 & 0.270 & 50.6 & 48.5 \\
    Ours  & 0.0170 & 0.0718 & 0.201 & 37.7 & 34.8 \\
    \hline
    \end{tabular}
}%
\caption{Quantitative inverse rendering results. Reflectance (albedo) errors are measured against multiview inverse rendering result \cite{kim2016multi} and normals against MVS results.}
\label{tab:am&nm_table}
\end{table}

\begin{figure}
\setlength{\tabcolsep}{1pt}
\renewcommand{\arraystretch}{1}
\centering
{
\begin{tabular}{ccc}
\includegraphics[width=2.7cm]{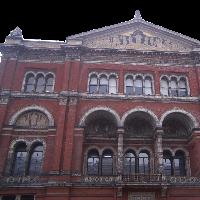}&
\includegraphics[width=2.7cm]{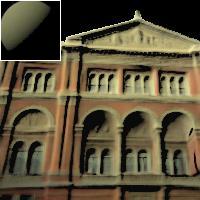}&
\includegraphics[width=2.7cm]{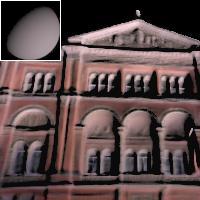}\\
Input & Relit 1 & Relit 2

\end{tabular}

    \vspace{-0.2cm}
    \caption{Relighting results from predicted albedo and normal maps (see Fig.~\ref{fig:teaser}, row 3). The novel lighting is shown in the upper left corner.}
    \label{fig:relit}
}

\end{figure}%

\mypara{Relighting}
Finally, as an example application we show that our inverse rendering result is sufficiently stable for realistic relighting. A scene from Fig.~\ref{fig:teaser} is relit in Fig.~\ref{fig:relit} with two novel illuminations. Both show realistic shading and overall colour balance.





\section{Conclusions}

We have shown for the first time that the task of inverse rendering can be learnt from real world images in uncontrolled conditions. Our results show that ``shape-from-shading'' in the wild is possible and are far superior to classical methods. It is interesting to ponder how this feat is achieved. We believe the reason this is possible is because of the large range of cues that the deep network can exploit, for example shading, texture, ambient occlusion, perhaps even high level semantic concepts learnt from the diverse data. For example, once a region is recognised as a ``window'', the possible shape and configuration is much restricted. Recognising a scene as a man-made building suggests the presence of many parallel and orthogonal planes. These sort of cues would be extremely difficult to exploit in hand-crafted solutions.

There are many promising ways in which this work can be extended. First, our modelling assumptions could be relaxed, for example using more general reflectance models and estimating global illumination effects such as shadowing. Second, our network could be combined with a depth prediction network. Either the two networks could be applied independently and then the depth and normal maps merged, or a unified network could be trained in which the normals computed from the depth map are used to compute the losses we use in this paper. Third, our network could benefit from losses used in training intrinsic image decomposition networks. For example, if we added the timelapse dataset of \cite{BigTimeLi18} to our training, we could incorporate their reflectance consistency loss to improve our albedo map estimates. Our code, trained model and inverse rendering benchmark data is available at $<$URL removed for review$>$. 
\newpage
{\small
\bibliographystyle{ieee}
\bibliography{egbib}
}

\end{document}